\documentclass[10pt,twocolumn,letterpaper]{article}

\usepackage{iccv}
\usepackage{times}
\usepackage{epsfig}
\usepackage{graphicx}
\usepackage{amsmath}
\usepackage{amssymb}

\usepackage{mathrsfs}
\usepackage{bbm}
\usepackage{booktabs}
\usepackage{multirow}
\usepackage{xcolor}
\usepackage{stfloats}
\usepackage[pagebackref=true,breaklinks=true,letterpaper=true,colorlinks,bookmarks=false]{hyperref}

\usepackage[accsupp]{axessibility}  

\iccvfinalcopy 

\def\iccvPaperID{0000} 

\ificcvfinal\fi

\begin{document}

\title{TCOVIS: Temporally Consistent Online Video Instance Segmentation}

\author{Junlong Li\ \ \
Bingyao Yu\ \ \
Yongming Rao\ \ \
Jie Zhou\ \ \
Jiwen Lu$\thanks{Corresponding author}$\\
{Department of Automation, Tsinghua University, China}\\
{Beijing National Research Center for Information Science and Technology, China}\\
}

\maketitle
\ificcvfinal\thispagestyle{empty}\fi

\begin{abstract}
   In recent years, significant progress has been made in video instance segmentation (VIS), with many offline and online methods achieving state-of-the-art performance. While offline methods have the advantage of producing temporally consistent predictions, they are not suitable for real-time scenarios. Conversely, online methods are more practical, but maintaining temporal consistency remains a challenging task. In this paper, we propose a novel online method for video instance segmentation, called TCOVIS, which fully exploits the temporal information in a video clip. The core of our method consists of a global instance assignment strategy and a spatio-temporal enhancement module, which improve the temporal consistency of the features from two aspects. Specifically, we perform global optimal matching between the predictions and ground truth across the whole video clip, and supervise the model with the global optimal objective. We also capture the spatial feature and aggregate it with the semantic feature between frames, thus realizing the spatio-temporal enhancement. We evaluate our method on four widely adopted VIS benchmarks, namely YouTube-VIS 2019/2021/2022 and OVIS, and achieve state-of-the-art performance on all benchmarks without bells-and-whistles. For instance, on YouTube-VIS 2021, TCOVIS achieves 49.5 AP and 61.3 AP with ResNet-50 and Swin-L backbones, respectively. Code is available at \url{https://github.com/jun-long-li/TCOVIS}.
\end{abstract}

\section{Introduction}
\label{sec:intro}
Video instance segmentation (VIS) is a challenging and representative video understanding task recently introduced in~\cite{yang2019video}. It aims at detecting, segmenting and tracking instances across a video. VIS is attracting increasing attention for various real-world applications such as video editing, video surveillance, augmented reality and autonomous driving. Recently introduced VIS methods can be roughly categorized into two groups: offline methods and online methods. Offline methods~\cite{bertasius2020classifying,hwang2021video,lin2021video,wang2021end,wu2022seqformer,wu2022efficient} take as input the whole video and perform the segmentation of instance sequence for the whole video at once. Online methods~\cite{fu2021compfeat,wu2022defense,ke2021prototypical,he2022inspro,yang2021crossover}, on the contrary, take as input a video frame by frame and generate the pre-frame object instances while associating the frame-wise results across frames. Both offline and online methods have achieved impressing performance on the VIS task.

\begin{figure}[t]
\begin{center}
\includegraphics[width=0.98\linewidth]{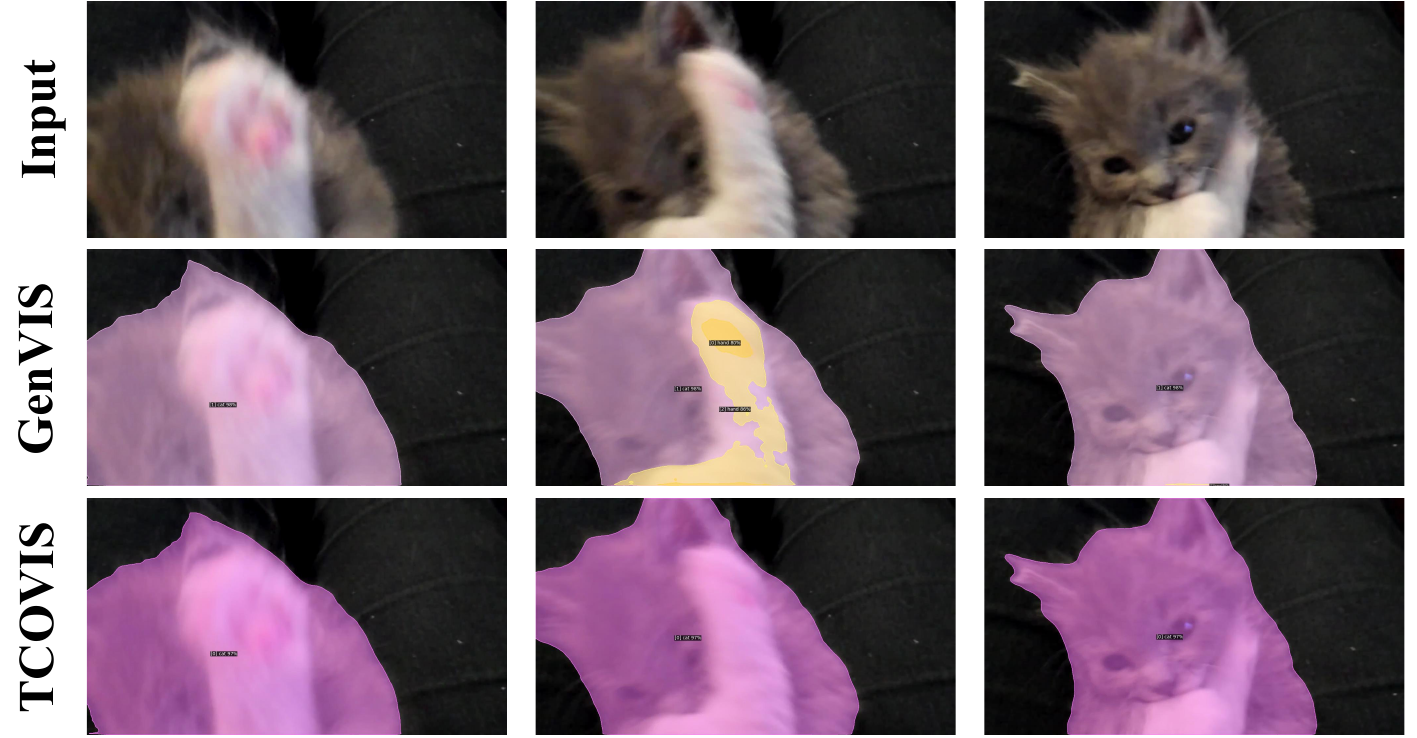}
\end{center}
\vspace{-2mm}
   \caption{Visualization of predictions from the previous online method (the online GenVIS~\cite{heo2022generalized}) and our TCOVIS. The previous method generates temporally inconsistent predictions, while our proposed TCOVIS achieves temporal consistency and outperforms the previous method (Best viewed in color).}
\label{fig:intro}
\vspace{-3mm}
\end{figure}

Offline methods have an inherent advantage in producing temporally consistent predictions, since delicate temporal communication and association mechanisms can be adopted throughout the video~\cite{yang2022temporally,wu2022seqformer,heo2022vita} to handle the overall temporal information and impose an explicit constraint on the temporal consistency. However, the video-in and video-out offline manner is not suitable for real-time scenarios. Conversely, online methods are more practical and making considerable progress but suffer from temporal inconsistency (as shown in Figure~\ref{fig:intro}), remaining a great challenge.

Online methods rely on specific instance association applied across frames, since only one frame is observed at a time. Existing association techniques can be grouped into two categories, including tracking-by-detection and query propagation-based paradigms. Tracking-by-detection methods~\cite{yang2019video,wu2022defense,wu2021track} generate the per-frame instances independently by existing instance segmentation models~\cite{he2017mask,tian2020conditional,cheng2022masked} and track instances via tracking heads~\cite{yang2019video} or instance embeddings matching~\cite{wu2022defense,huang2022minvis}. In this way, the features of different  frames are isolated before tracking, which results in temporal inconsistency. Query propagation-based methods~\cite{heo2022generalized,he2022inspro,zhan2022robust} are inspired by query-based methods~\cite{carion2020end,sun2021sparse} and they propagate the query across  frames to decode a unique instance without heuristic matching algorithms. Despite the explicit temporal link of queries, the temporal consistency is impaired by the Local Matching and Propagating (LocPro) scheme, where they first perform local optimal matching between the predictions and ground truth at the beginning of the video, and then propagate the assignment across frames, forcing all features from subsequent frames to follow. The LocPro is not suitable for the holistic optimization across frames and results in temporal inconsistency. Thus, achieving temporal consistency is challenging for online methods and also not comprehensively investigated by previous online VIS methods. 

In this paper, we propose a novel online method for video instance segmentation, named TCOVIS, to fully exploit the temporal information within a video. We take as the baseline framework an existing online VIS model (GenVIS~\cite{heo2022generalized}) with the query propagation-based instance association. We first introduce the global instance assignment strategy to perform global optimal matching. Different from the previous online methods~\cite{heo2022generalized,zhan2022robust,he2022inspro}, which obtain per-frame matching cost, assign labels locally on the beginning frame and propagate across frames, we collect the predictions across frames, compute the global matching cost with the video segmentation ground truth and supervise the model with the global instance assignment, encouraging features across the video to be optimized for a global optimal objective. As online methods focus on improving the representative ability of the semantic instance embeddings~\cite{yang2021crossover,wu2022defense}, which is achieved via learning more discriminative semantic embedding in those using heuristic matching~\cite{wu2022defense,huang2022minvis}, or via reviewing the memory of semantic embedding across frames~\cite{heo2022generalized} in query propagation-based methods, the spatial features are not comprehensively investigated. We further propose spatio-temporal enhancement module, leveraging the spatial information from the previous frame to enhance the temporal consistency. We perform spatial matting on the pixel embeddings to retrieve the instance-wise spatial features and adopt the cross-attention layer to aggregate the spatial and semantic features across frames, thus realizing the spatio-temporal enhancement. As shown in Figure~\ref{fig:intro}, the previous online method~\cite{heo2022generalized}
produces temporally inconsistent results, as exemplified by an object abruptly appearing in front of the cat in a mid-frame, while the proposed 
TCOVIS outperforms the previous one and generates temporally consistent predictions.

To validate the effectiveness of the proposed method, experiments are conducted on four widely adopted VIS benchmarks, i.e., YouTube-VIS 2019~\cite{yang2019video}, YouTube-VIS 2021, YouTube-VIS 2022 and Occluded VIS (OVIS)~\cite{qi2022occluded}. Without bells-and-whistles, our proposed method achieves state-of-the-art performance on all benchmarks, outperforming other online methods, e.g., on YouTube-VIS 2021, TCOVIS achieves 49.5 AP and 61.3 AP with ResNet-50 and Swin-L backbones, respectively.

Our main contributions are summarized as follows:
\begin{itemize}
    \item TCOVIS performs a novel global instance assignment strategy for online video instance segmentation. The model is optimized for the global optimal objective to generate more temporally consistent predictions.
    \item The further proposed spatio-temporal enhancement module captures the spatial feature and aggregates it with the semantic feature between frames, which fully utilizes the spatial information and facilitates the temporal consistency enhancement.
    \item The proposed method achieves state-of-the-art performance on four widely used video instance segmentation benchmarks (YouTube-VIS 2019/2021/2022 and OVIS). Such achievements demonstrate the effectiveness of our proposed method. 
\end{itemize}

\section{Related Works}

\noindent\textbf{Offline Video Instance Segmentation.}
Offline methods take as input the whole video and predict instance sequence for all frames at once. Mask propagation and box ensemble techniques are used to improve the predictions and association~\cite{athar2020stem,bertasius2020classifying,lin2021video}, but they are not end-to-end learnable due to the complex inference process. VisTR~\cite{wang2021end} extends DETR~\cite{zhu2021deformable} from the image domain and introduces the transformer~\cite{vaswani2017attention} to the VIS domain. EfficientVIS~\cite{wu2022efficient} and IFC~\cite{hwang2021video} relax the heavy overhead of VisTR via an iterative query-video interaction and memory token communication, respectively. TeViT~\cite{yang2022temporally} contains a vision transformer~\cite{dosovitskiy2021an} backbone instead of CNN and efficiently builds correspondence between the instance and query. VITA~\cite{heo2022vita} models relationships among instances with the distilled condensed object tokens, without using the dense spatio-temporal backbone features. Offline methods exploit rich temporal knowledge from the whole clip and have the advantage of producing temporally consistent results, however, the offline manner is not suitable for the application in real-time scenarios.

\noindent\textbf{Online Video Instance Segmentation.} 
Instead of processing the entire video before predictions, online methods only leverage the information from the previous frames and segment the video frame-by-frame. The association paradigms of the previous online methods roughly fall into two groups: Tracking-by-detection and Query-propagation.

Most online methods~\cite{yang2019video,wang2021end2,liu2021sg,wu2022defense} follow the tracking-by-detection paradigm. MaskTrack R-CNN~\cite{yang2019video} is the baseline method and extends the Mask R-CNN~\cite{he2017mask} with an extra tracking head for temporal association. CrossVIS~\cite{yang2021crossover} proposes a crossover learning scheme to utilize the current contextual information for other frames. VISOLO~\cite{han2022visolo} builds on the image instance segmentation method SOLO~\cite{wang2020solo} and takes advantage of the grid form previous information for memory matching and features aggregation. MinVIS~\cite{huang2022minvis} and IDOL~\cite{wu2022defense} make use of the discriminative instance embeddings for matching between frames. With the heuristic matching technique designed for instance association across frames, temporal inconsistency comes from the frame-wise modeling before tracking.

Object association with query propagation has been explored in the multi-object tracking (MOT) task~\cite{meinhardt2022trackformer,zeng2022motr}. TrackFormer~\cite{meinhardt2022trackformer} tracks the seen objects of previous frames with a track query subset and detects the newly appeared objects in current frame with an extra object query subset. MOTR~\cite{zeng2022motr} extends the paired-frames training scheme to multiple frames for long-range temporal association. The query propagation-based object association is recently introduced to VIS~\cite{zhan2022robust,he2022inspro,heo2022generalized}. ROVIS~\cite{zhan2022robust} follows TrackFormer~\cite{meinhardt2022trackformer} to detect and track instances with two subsets of queries. InsPro~\cite{he2022inspro} and GenVIS~\cite{heo2022generalized} propagate the queries without heuristics, i.e., handcrafted rules to combine two types of queries, and achieve association across frames. However, previous query propagation-based methods also propagate the local assignment to supervise the model with the local optimal results, which leads to temporal inconsistency across the entire video. We adopt the query propagation framework but introduce the global instance assignment to enhance the temporal consistency.

\begin{figure*}
\begin{center}
\includegraphics[width=0.95\textwidth]{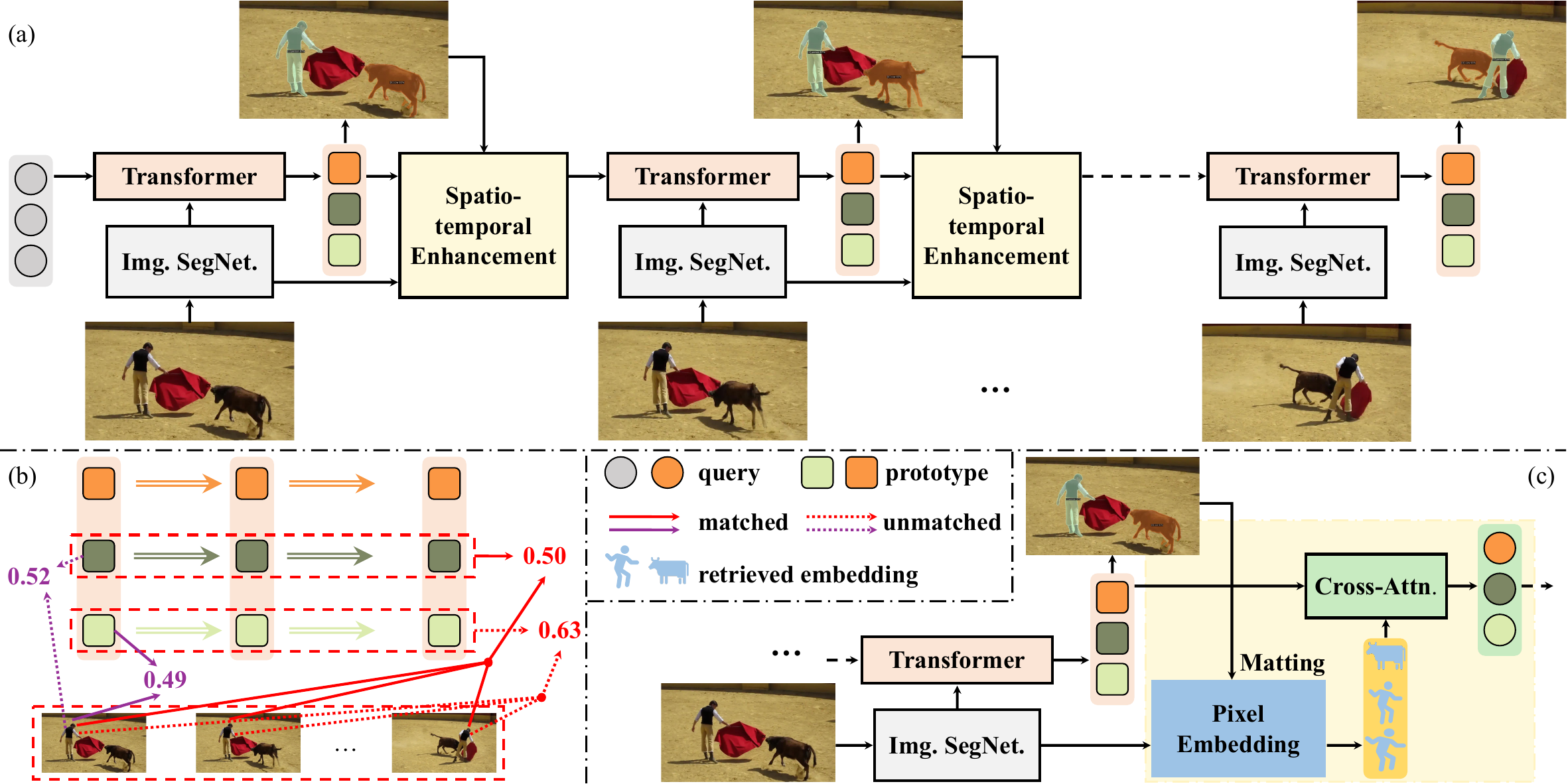}
\end{center}
\vspace{-2mm}
   \caption{(a) The overall illustration of our TCOVIS. It supervises the model with a global optimal objective during training and utilizes spatial features via the enhancement module between frames. (b) Details of global instance assignment. Different from the local matching and propagating technique, we conduct global optimal matching and assignment. (c) In the spatio-temporal enhancement module, we perform matting on pixel embeddings according to predicted masks and then aggregate the spatial and semantic features between frames to enhance temporal consistency. Numbers in colors denote the \textcolor{violet}{local matching loss} and \textcolor{red}{global matching loss} (Best viewed in color).}
\label{fig:method}
\vspace{-3mm}
\end{figure*}

\section{Method}
Given a video clip with consecutive image frames, online video instance segmentation methods generate frame-level object instances upon on instance segmentation models~\cite{tian2020conditional,cheng2022masked}, utilizing the instance queries propagated from previous frames~\cite{wu2022efficient,heo2022generalized}. We have already discussed that better performance of the online VIS method relies on more temporally consistent instance features among a video. To this end, we propose a novel end-to-end online VIS method TCOVIS (Figure~\ref{fig:method}), to improve the temporal consistency of instance features via the global instance assignment strategy and spatio-temporal enhancement module. In this section, we first introduce the online VIS pipeline in Section~\ref{sec:online_vis}. Then the details of the proposed global instance assignment strategy and spatio-temporal enhancement module will be described in Section~\ref{sec:gia} and Section~\ref{sec:ste}, respectively. Finally, in Section~\ref{sec:loss}, we describe the overall loss
for training the model end-to-end.

\subsection{Online Video Instance Segmentation}
\label{sec:online_vis}
Following state-of-the-art VIS methods~\cite{huang2022minvis,cheng2021mask2former}, we adopt the advanced Masked-attention Mask Transformer (Mask2Former~\cite{cheng2022masked}) as the image instance segmentation network (Img. SegNet.) in this paper. 
Assume that the input video clip with T frames is denoted as $x \in \mathbb{R}^{T \times 3 \times H \times W}$. With each frame $H \times W$ as input, frame-wise activation map is extracted by the backbone and Transformer encoder. Then following the query-based mechanism of DETR~\cite{carion2020end}, $N_{fq}$ object queries of $C$ dimensions are used to parse an input frame, which are called frame object queries $f \in \mathbb{R}^{C \times N_{fq}}$. Each object in the frame is decoded by the frame object queries from the spatial features through a multiple-level Transformer decoder, and represented as an object embedding of $C$ dimensions. The object embeddings are used for classification and together with the pixel embeddings from the pixel decoder generating the mask for object instances. Class predictions are produced through a linear layer from the object embeddings. Mask embeddings are generated by an MLP linked to the object embeddings, and the model finally segments objects by pixel-wise dot product between per-pixel embeddings $\mathcal{P} \in \mathbb{R}^{C \times \frac{H}{S} \times \frac{W}{S}}$ and mask embeddings $\mathcal{M} \in \mathbb{R}^{C \times N_{fq}}$, where $S$ is the stride of the spatial feature map.

As for online video segmentation, we follow the online scheme of GenVIS~\cite{heo2022generalized} which propagates the instance queries from previous frames. To resolve the computational limitation, the framework adopts VITA~\cite{heo2022vita} that regards the frame object queries $f$ as a concise representation of objects in a frame and then feeds them into the \emph{Object Encoder} $\mathscr{E}$ for intra-frame relationship. The \emph{Object Decoder} $\mathscr{D}$ takes as input $N_v$ video instance queries $q$ and aggregates information from frame object queries. The temporal instance association is implemented through a query-based temporal propagation mechanism, where the output of $\mathscr{D}$, instance prototypes denoted as $p$, are concise representations of instances~\cite{heo2022generalized}, are not only utilized for classifying and segmenting in current frame at $t$, but also serve as the instance queries for the next frame at $t+1$, i.e.,
\begin{equation}\label{eq:propagation}
  q^{t+1} = p^{t} = \mathscr{D}(q^{t}, \mathscr{E}(f^{t})).
\end{equation} 
With this propagation mechanism, the model can simply run in an online manner and associate the frame-wise outputs without heuristic matching algorithms.

 We leave out the Instance Prototype Memory module in ~\cite{heo2022generalized}. During the training process, we freeze the image instance segmentation model and only train the following modules, to efficiently make use of memory. More implementation details are described in Section~\ref{sec:imple}. 

 \subsection{Global Instance Assignment}
 \label{sec:gia}
With the query propagation mechanism discussed above, the associated video instance queries along the temporal dimension $\{q^{1:T}_k\}_{1:N_v}$, are used to extract the features of a unique instance, i.e. the $k$-th instance, throughout the whole input video clip. Meanwhile, the associated instance prototypes $\{p^{1:T}_k\}_{1:N_v}$, i.e. the corresponding outputs 
from $\mathscr{D}$, represent this unique instance within the whole video. For brevity, we define the associated instance prototypes/features as a \textbf{set} of prototypes/features.

Previously, online VIS methods that associate frame-level results without heuristic matching algorithms~\cite{heo2022generalized,zhan2022robust,he2022inspro} adopt a \emph{Local Matching and Propagating (LocPro)} technique. This technique matches the predictions and ground truth on a frame level during training. In other words, it conducts one-to-one bipartite matching between ground truth instances and predictions of a frame (what we regard as \emph{Local Matching}), and propagates the matching from previous frames to subsequent frames. Specifically, given a video clip consisting of $T$ consecutive frames, it first computes the pair-wise matching cost $\mathcal{L}_{match}^t$ between predictions and ground truth instances on the initial frame (the first frame or the frame objects appear):
\begin{equation}\label{eq:frame_matching}
  \mathcal{L}_{match}^{t=1} = \lambda_{cls} \mathcal{L}_{cls}^{t=1} + \lambda_{bce} \mathcal{L}_{bce}^{t=1} + \lambda_{dice} \mathcal{L}_{dice}^{t=1}.
\end{equation} 
The frame-wise $\mathcal{L}_{match}^{t}$ is composed of categorical loss and mask loss. The categorical loss adopts the cross entropy loss $\mathcal{L}_{cls}^{t}$. The mask loss consists of a binary cross entropy loss $\mathcal{L}_{bce}^{t}$ and a dice loss~\cite{milletari2016v}. From the cost matrix, it follows DETR~\cite{carion2020end} and uses Hungarian algorithm~\cite{kuhn1955hungarian} for optimal frame-level matching. 

As discussed in Section~\ref{sec:intro}, \emph{Local Matching} only obtains optimal matching on the initial frame, and cannot achieve the global optimal matching of the whole video. As illustrated in Figure~\ref{fig:method} (b), both dark and light green prototypes attempt to represent the person. With local matching, the ground truth will be assigned to the light one with a smaller local matching cost, however, the dark one performs better from a global perspective and is supposed to be assigned. Training the model with \emph{LocPro} forces all the predictions of subsequent frames to conform to the initial frame, resulting in temporal inconsistency of the instance features among the video, because inappropriate previous matching brings accumulative error to the model.

We introduce \textbf{Global Instance Assignment} (GIA) strategy and expect two functionalities: (1) all frames in the video clip are considered when conducting global matching, and (2) the global assignment encourages the instance features among frames to be optimized for a global optimal objective, both of which serve the temporal consistency of instance features.

During training, different from previous online VIS methods~\cite{fu2021compfeat,wu2022defense} that conduct frame-level local matching and provide supervision signals to each frame according to the matched pairs, we leave out the halfway matching and supervision. Consecutive input frames of a video clip pass through the VIS model in an online manner and the model generates the predictions of all frames $\{\hat{\mathbf{y}}_k\}_{1:N_v} = \{\hat{y}_k^t\}^{1:T}_{1:N_v}$, each of which consists of a category probability $\hat{c}_k^t$ and a segmentation mask probability $\hat{m}_k^t$. To conduct global assignment, we collect predictions of all frames and as well the ground truth video instance segmentation $\{\mathbf{y}_k\}_{1:N_{gt}} = \{y_k^{1:T}\}_{1:N_{gt}}$, including the category label $c_k$ and its binary segmentation masks $\mathbf{m}_k = m_k^{1:T}$. Since the ground truth of an instance only has one category label, we first compute the average predicted category probability across a video clip: $\Bar{c}_k = \sum_{t=1}^{T} \hat{c}_k^t / T$ and also collect the masks $\hat{\mathbf{m}}_k = \hat{m}_k^{1:T}$. The global matching cost is defined as:
\begin{equation}\label{eq:global_match}
\begin{aligned}
  \mathcal{L}_{match}^{global} &= \lambda_{cls} \mathcal{L}_{cls}(c_k, \Bar{c}_{\sigma (k)}) + \lambda_{bce} \mathcal{L}_{bce}(\mathbf{m_k}, \hat{\mathbf{m}}_{\sigma (k)}) \\
  &+ \lambda_{dice} \mathcal{L}_{dice}(\mathbf{m_k}, \hat{\mathbf{m}}_{\sigma (k)}),
\end{aligned}
\end{equation} 
where $\sigma \in \mathfrak{S}_{N_v}$ is a permutation of $N_v$ elements. One-to-one bipartite global matching between $\{\hat{\mathbf{y}}_k\}_{1:N_v}$ and $\{\mathbf{y}_k\}_{1:N_{gt}}$ is performed to find the global optimal assignment and the objective can be formally described as:
\begin{equation}\label{eq:assign}
\hat{\sigma} = \underset{\sigma \in \mathfrak{S}_{N_v}}{\arg\max} \sum_{k=1}^{N_{gt}}\mathcal{L}_{match}^{global}(\mathbf{y}_k,\hat{\mathbf{y}}_{\sigma(k)}).
\end{equation} 
Following prior work~\cite{wang2021max,fang2021instances,zhu2021deformable}, we use Hungarian algorithm~\cite{kuhn1955hungarian} to search for the global optimal assignment. In contrast to the prior methods that compute matching loss only for $t=1$, our method considers the masks and ground truth of the whole clip, computes the matching loss globally, and conducts the global assignment. Finally, given the global optimal assignment, we use the $N_{gt}$ matched video-level predictions to supervise the model with the globally matched instances across the entire video. 

With the proposed assignment strategy, GIA, the temporal consistency of instance features across the video clip can be effectively enhanced, since we consider all frames as a whole to search for the optimal objective. Specifically, when the set of instance features across the video represents the target instance well in the first few frames, but fails to track it later, this strategy helps to find a more appropriate set to be optimized. As the global matching cost is lower, the selected features fit the target more closely and are more temporally consistent.

 \subsection{Spatio-temporal Enhancement}
 \label{sec:ste}
  Previous online video instance segmentation methods focus on improving the representative ability of the semantic instance embeddings~\cite{yang2021crossover,wu2022defense,huang2022minvis,heo2022generalized}. The spatial features are not comprehensively investigated to boost the temporal association for online VIS. Thus, we further introduce \textbf{Spatio-temporal Enhancement} module (STE), leveraging the spatial information from the previous frame to enhance the temporal consistency of the online model.

 Given the $t$-th frame in the video clip, with the framework described in Section~\ref{sec:online_vis}, the mask embedding of the $k$-th instance is generated from the instance prototype through an MLP: $\mathcal{M}_k^t = MLP(p_k^t)$, and then the model segments the mask $\hat{m}_k^t$ by pixel-wise dot product between $\mathcal{P}_k^t$ and $\mathcal{M}_k^t$, which can be formulated as: $\hat{m}_{k,i,j}^t = \langle \mathcal{P}_{k,i,j}^t, \mathcal{M}_k^t \rangle$, where $i$ and $j$ denote the spatial position of the pixel. 

 As illustrated in Figure~\ref{fig:method} (c), the spatial information of the frame is encoded in the pixel embeddings. To extract instance-wise spatial features, they can be exploited together with the predicted mask. Specifically, we perform spatial matting on pixel embeddings $\mathcal{P}_k^t$, similar to image matting, to retrieve the instance-wise pixel embedding according to $\hat{m}_k^t$. For each instance, we conduct pixel-wise multiplication between the original pixel embeddings and the binary mask to obtain the retrieved embeddings: 
 \begin{equation}\label{eq:retrieve}
  \mathcal{R}_k^t = \mathcal{P}_k^t \odot \hat{m}_k^t,
\end{equation}
where $\odot$ denotes the element-wise multiplication. Many of the retrieved embeddings $\mathcal{R}_k^t$ are redundant since they describe the same instance, and directly mining the spatial information with them is computationally inefficient. Average pooling is adopted to obtain a concise representation of the spatial features for each instance: 
 \begin{equation}\label{eq:avg_pool}
  \mathcal{S}_k^t = \frac{\sum_{i,j} \mathcal{R}_{k,i,j}^t}{\sum_{i,j}\mathbbm{1}(\hat{m}_{k,i,j}^t = 1)},
\end{equation}
where $\mathbbm{1}(\cdot)$ is the indicator function. The concise spatial features $\mathcal{S}_k^t$ are sent to the next frame for temporal association.

Having received from the previous frame the propagated semantic instance queries, i.e., $q^{t+1} = p^t$, and the spatial features, we follow the standard multi-head cross-attention layer (MHCA)~\cite{vaswani2017attention} to incorporate the features from two aspects. The instance prototype $p^{t}$ is used as the query to decode the spatial features:
 \begin{equation}\label{eq:mhca}
  {q}_k^{t+1} = \mathrm{MHCA}(p_k^{t}, \mathcal{S}_{1:N_v}^t),
\end{equation}
where $q_{k}^{t+1}$ is the updated instance query now. Notably, the positional embedding is shared by the spatial feature and instance query with regard to the same instance, which helps the model align the instance-wise information between frames. Finally, the updated instance query is fed into the \emph{Object Decoder} $\mathscr{D}$ to produce spatio-temporally enhanced features.

By performing the proposed enhancement module on the spatial features between frames, we effectively boost the spatio-temporal association of the features. In this way, the temporal consistency of the features is enhanced via the information from the spatial dimension, and 
the online video instance segmentation model manages to predict more temporally consistent results. 

\subsection{Overall Loss}
\label{sec:loss}
The overall loss for training with a video clip as input is a linear combination of categorical and mask losses using the one-to-one global optimal assignment $\hat{\sigma}$:
\begin{equation}\label{eq:loss}
  \begin{aligned}
  \mathcal{L}_{overall} &= \lambda_{cls} \mathcal{L}_{cls}(c_k, \Bar{c}_{\hat{\sigma} (k)}) + \lambda_{bce} \mathcal{L}_{bce}(\mathbf{m_k}, \hat{\mathbf{m}}_{\hat{\sigma} (k)}) \\
  &+ \lambda_{dice} \mathcal{L}_{dice}(\mathbf{m_k}, \hat{\mathbf{m}}_{\hat{\sigma} (k)}),
\end{aligned}
\end{equation}
where $\mathcal{L}_{cls}$ is the cross entropy loss, $\mathcal{L}_{bce}$ is the binary cross entropy loss and $\mathcal{L}_{dice}$ is the dice loss~\cite{milletari2016v}.

\section{Experiments}
In this section, we evaluated the proposed TCOVIS on four benchmark datasets. 
Furthermore, we provided in-depth ablation studies of the effectiveness of TCOVIS.
Finally, we presented several visualizations of the predictions from our model.

\subsection{Datasets}
We evaluated our approach on four VIS datasets:
YouTube-VIS 2019 dataset~\cite{yang2019video}, YouTube-VIS 2021 dataset~\cite{yang2019video}, YouTube-VIS 2022 dataset~\cite{yang2019video} and OVIS dataset~\cite{qi2022occluded}. We present a brief description of them:

\textbf{YouTube-VIS 2019 \& 2021 \& 2022:} YouTube-VIS 2019~\cite{yang2019video} is the first VIS dataset and comprises 40 predefined categories of objects. 
This dataset included 4,883 unique video instances with 131,000 high-quality manual annotations. We followed the widely utilized
training/test set split: 2,238 videos were selected for training, 302 videos were adopted for validation and 343 videos were used for testing. Further, YouTube-VIS 2021 improved the 40-category label set and added 4883 more unique video instances. Then, YouTube-VIS 2022 contained 71 additional long evaluation videos on the top of YouTube-VIS 2021.

\textbf{OVIS:} 
Occluded video instance segmentation (OVIS) is also a challenging VIS dataset. OVIS included 901 videos in total with 25 semantic categories. 
This dataset contained 5,223  unique video instances and we followed the widely utilized
training/test set split: 607 videos were selected for training, 607 videos were used for validation and 154 videos were adopted for testing.

Following~\cite{yang2019video}, the video-level average precision (AP) and average recall (AR) were adopted as the evaluation metrics on both YouTube-VIS and OVIS.

\begin{table*}
\centering
{
\begin{tabular}{@{}c|lc|c|ccccc|ccccc@{}}
\hline
\multicolumn{3}{c|}{\multirow{2}{*}{Method}}                        & \multirow{2}{*}{Type} & \multicolumn{5}{c|}{YouTube-VIS 2019} & \multicolumn{5}{c}{YouTube-VIS 2021}\\
\multicolumn{3}{l|}{}                                               &         & AP        & AP$_{50}$ & AP$_{75}$ & AR$_1$    & AR$_{10}$ & AP        & AP$_{50}$ & AP$_{75}$ & AR$_1$    & AR$_{10}$ \\
    \hline
    \hline
    
    \multirow{13}{*}{\rotatebox{90}{ResNet-50}}
    
    & \multicolumn{2}{l|}{EfficientVIS~\cite{wu2022efficient}}         & \textcolor{lightgray}{Offline}     & 37.9      & 59.7      & 43.0      & 40.3      & 46.6 
                                                                    & 34.0      & 57.5      & 37.3      & 33.8      & 42.5  \\
    & \multicolumn{2}{l|}{IFC~\cite{hwang2021video}}                           & \textcolor{lightgray}{Offline}     & 41.2      & 65.1      & 44.6      & 42.3      & 49.6                                                                   & 35.2      & 55.9      & 37.7      & 32.6      & 42.9  \\
    & \multicolumn{2}{l|}{Mask2Former-VIS~\cite{cheng2021mask2former}}   & \textcolor{lightgray}{Offline}     & 46.4      & 68.0      & 50.0      & -         & -  & 40.6      & 60.9      & 41.8      & -         & -     \\
    & \multicolumn{2}{l|}{TeViT$^\dagger$~\cite{yang2022temporally}}                       & \textcolor{lightgray}{Offline}      & 46.6      & 71.3      & 51.6      & 44.9      & 54.3   & 37.9      & 61.2      & 42.1      & 35.1      & 44.6  \\
    & \multicolumn{2}{l|}{SeqFormer~\cite{wu2022seqformer}}               & \textcolor{lightgray}{Offline}    & 47.4      & 69.8      & 51.8      & 45.5      & 54.8                                                                   & 40.5      & 62.4      & 43.7      & 36.1      & 48.1  \\
     
    & \multicolumn{2}{l|}{VITA~\cite{heo2022vita}}                         & \textcolor{lightgray}{Offline}     & 49.8      & 72.6     & 54.5      & 49.4      & 61.0     
                                                                                    & 45.7      & 67.4      & 49.5      & 40.9      & 53.6  \\
    & \multicolumn{2}{l|}{\text{GenVIS}$_{\text{semi-online}}$~\cite{heo2022generalized}}           & \textcolor{lightgray}{Offline}     & 51.3  & 72.0    & 57.8  & 49.5  & 60.0  
                                                                                    & 46.3  & 67.0 & 50.2 & 40.6 & 53.2 \\     
\cmidrule{2-14}
  & \multicolumn{2}{l|}{CrossVIS~\cite{yang2021crossover}}                 & Online     & 36.3      & 56.8      & 38.9      & 35.6      & 40.7      
                                                                                    & 34.2      & 54.4      & 37.9      & 30.4      & 38.2  \\
    & \multicolumn{2}{l|}{VISOLO~\cite{han2022visolo}}                     & Online     & 38.6      & 56.3      & 43.7      & 35.7      & 42.5      
                                                                                    & 36.9      & 54.7      & 40.2      & 30.6      & 40.9  \\
    & \multicolumn{2}{l|}{MinVIS~\cite{huang2022minvis}}                     & Online     & 47.4      & 69.0      & 52.1      & 45.7      & 55.7      
                                                                                    & 44.2      & 66.0      & 48.1      & 39.2      & 51.7  \\
    & \multicolumn{2}{l|}{IDOL~\cite{wu2022defense}}                         & Online    & 49.5      & \textbf{74.0}     & 52.9      & 47.7      & 58.7   
                                                                                    & 43.9      & \underline{68.0}      & 49.6      & 38.0      & 50.9  \\
    & \multicolumn{2}{l|}{\text{GenVIS}$_{\text{online}}$~\cite{heo2022generalized}}                & Online & \underline{50.0} & 71.5 & \underline{54.6} & \underline{49.5} & \underline{59.7}
                                                                                    & \underline{47.1} & 67.5 & \underline{51.5} & \textbf{41.6} & \underline{54.7} \\
    & \multicolumn{2}{l|}{\textbf{TCOVIS}}           & Online     & \textbf{52.3}& \underline{73.5}	& \textbf{57.6}	& \textbf{49.8}	& \textbf{60.2} & \textbf{49.5}	& \textbf{71.2}	& \textbf{53.8}	& \underline{41.3}	& \textbf{55.9}\\                              
    \hline
    \hline
    
    \multirow{8}{*}{\rotatebox{90}{Swin-L}}
    & \multicolumn{2}{l|}{SeqFormer~\cite{wu2022seqformer}}               & \textcolor{lightgray}{Offline}        & 59.3      & 82.1      & 66.4      & 51.7      & 64.4  
                                                                    & 51.8      & 74.6      & 58.2      & 42.8      & 58.1  \\
    & \multicolumn{2}{l|}{Mask2Former-VIS~\cite{cheng2021mask2former}}   & \textcolor{lightgray}{Offline}        & 60.4      & 84.4      & 67.0      & -         & -         
                                                                & 52.6      & 76.4      & 57.2      & -         & -     \\
    & \multicolumn{2}{l|}{VITA~\cite{heo2022vita}}                         & \textcolor{lightgray}{Offline}        & 63.0      & 86.9      & 67.9      & 56.3      & 68.1
                                                                                    & 57.5      & 80.6   & 61.0      & 47.7   & 62.6  \\   
    & \multicolumn{2}{l|}{\text{GenVIS}$_{\text{semi-online}}$~\cite{heo2022generalized}}           & \textcolor{lightgray}{Offline}        & 63.8     & 85.7              & 68.5    & 56.3     & 68.4
                                                                                    & 60.1     & 80.9     & 66.5     & 49.1     & 64.7\\

    \cmidrule{2-14}
    & \multicolumn{2}{l|}{MinVIS~\cite{huang2022minvis}}                     & Online        & 61.6      & 83.3      & 68.6      & 54.8      & 66.6 
                                                                                    & 55.3      & 76.6      & 62.0      & 45.9      & 60.8  \\   
    & \multicolumn{2}{l|}{IDOL~\cite{wu2022defense}}                         & Online        & \textbf{64.3}      & \textbf{87.5}      & \textbf{71.0}      & 55.6      & \underline{69.1}
                                                                                    & 56.1      & 80.8      & 63.5      &  45.0      & 60.1  \\
    & \multicolumn{2}{l|}{\text{GenVIS}$_{\text{online}}$~\cite{heo2022generalized}}                & Online      & 64.0 & 84.9 & 68.3 & \textbf{56.1} & \textbf{69.4}  
                                                                                    & \underline{59.6}  & \underline{80.9} & \underline{65.8} & \textbf{48.7} & \underline{65.0} \\
     & \multicolumn{2}{l|}{\textbf{\text{TCOVIS}}}                & Online     &\underline{64.1}	&\underline{86.6}	&\underline{69.5}&\underline{55.8}	&69.0
     &\textbf{61.3}	&\textbf{82.9}	&\textbf{68.0}	&\underline{48.6}	&\textbf{65.1}\\  
    \hline
    \end{tabular}
}
\vspace{2mm}
\caption{
Quantitative results on \textbf{YouTube-VIS 2019 and 2021 validation} sets.
The results are respectively grouped by method types (Offline or Online) and backbone networks (ResNet-50 and Swin-L). 
We \textbf{bold} the best performance and \underline{underline} the second. $\dagger$ denotes using MsgShifT~\cite{yang2022temporally} backbone which has a similar weight scale with ResNet-50.
}
\vspace{-2mm}
\label{tab:ytvis2019_2021}
\end{table*}

\subsection{Implementation Details}
\label{sec:imple}
We adopted the framework of GenVIS~\cite{heo2022generalized} which is built on VITA~\cite{heo2022vita}, but left out the similarity loss in VITA and the memory module in GenVIS. With the global assignment, the total loss as well the hyper-parameters were set the same as the video-level loss in VITA, which is a temporally extended loss function~\cite{hwang2021video}. The model was trained with pseudo-videos from COCO images~\cite{lin2014microsoft} as data augmentation, and with a batch size of 8 video clips of 6 frames. As we froze the backbone and image segmentation model, all experiments were conducted with 8 RTX 2080 Ti GPUs. The method was implemented on detectron2~\cite{detectron2}.

\subsection{Main Results}
Following the standard evaluation metrics~\cite{yang2019video}, we compared TCOVIS with state-of-the-art approaches on four VIS benchmarks: YouTube-VIS 2019/2021/2022 and OVIS. 

\begin{table}
\centering
\resizebox{\linewidth}{!}
{
\begin{tabular}{@{}c|ll|c|ccccc@{}}

\hline
\multicolumn{3}{c|}{Method} &Type    & AP        & AP$_{50}$ & AP$_{75}$ & AR$_{1}$  & AR$_{10}$ \\
\hline
\hline
\multirow{5}{*}{\rotatebox{90}{ResNet-50\space\space}}

& \multicolumn{2}{l|}{VITA~\cite{heo2022vita}}        &\textcolor{lightgray}{Offline}      & 32.6      & 53.9      & 39.3      & 30.3      & 42.6 \\
& \multicolumn{2}{l|}{\text{GenVIS}~\cite{heo2022generalized}} &\textcolor{lightgray}{Offline} & 37.2      & 58.5      & 42.9      & 33.2      & 40.4 \\
\cmidrule{2-9}
& \multicolumn{2}{l|}{MinVIS~\cite{huang2022minvis}}   &Online       & 23.3      & 47.9      & 19.3      & 20.2      & 28.0 \\
& \multicolumn{2}{l|}{\text{GenVIS}~\cite{heo2022generalized}} &Online& \underline{37.5}     & \textbf{61.6}      & \underline{41.5}      & \underline{32.6}      & \underline{42.2} \\

& \multicolumn{2}{l|}{\text{TCOVIS}} &Online& \textbf{38.6}	& \underline{59.4}	& \textbf{41.6}	& \textbf{32.8}	& \textbf{46.7} \\
\hline

\multirow{5}{*}{\rotatebox{90}{Swin-L\space}}

& \multicolumn{2}{l|}{VITA$^\star$~\cite{heo2022vita}}    &\textcolor{lightgray}{Offline}         & 41.1      & 63.0      & 44.0      & 39.3     & 44.3 \\
& \multicolumn{2}{l|}{$\text{GenVIS}$ ~\cite{heo2022generalized}} &\textcolor{lightgray}{Offline}& 44.3 & 69.9 & 44.9 & 39.9 & 48.4 \\
\cmidrule{2-9}
& \multicolumn{2}{l|}{MinVIS$^\star$~\cite{huang2022minvis}}   &Online      & 33.1      & 54.8      & 33.7      & 29.5     & 36.6 \\
& \multicolumn{2}{l|}{$\text{GenVIS}$~\cite{heo2022generalized}}&Online & \underline{45.1} & \underline{69.1} & \underline{47.3} & \underline{39.8} & \underline{48.5} \\

& \multicolumn{2}{l|}{\text{TCOVIS}} &Online& \textbf{51.0}	& \textbf{73.0}	& \textbf{53.5}	& \textbf{41.7}	& \textbf{56.5}\\
\hline
\end{tabular}
}
\vspace{1mm}
\caption{
Quantitative results on \textbf{YouTube-VIS 2022 validation} dataset.
We \textbf{bold} the highest accuracy and \underline{underline} the second.
$\star$: Reproduced by~\cite{heo2022generalized}.
}
\vspace{-4mm}
\label{tab:ytvis_2022}
\end{table}

\begin{table}
\centering
\resizebox{\linewidth}{!}
{
\begin{tabular}{@{}c|cc|c|ccccc@{}}

\hline
\multicolumn{3}{l|}{Method}&Type & AP        & AP$_{50}$     & AP$_{75}$     & AR$_{1}$      & AR$_{10}$ \\
\hline
\hline

\multirow{9}{*}{\rotatebox{90}{ResNet-50}}

& \multicolumn{2}{l|}{TeViT$^\dagger$~\cite{yang2022temporally}}&\textcolor{lightgray}{Offline} & 17.4      & 34.9          & 15.0          & 11.2          & 21.8  \\
& \multicolumn{2}{l|}{VITA~\cite{heo2022vita}}    &\textcolor{lightgray}{Offline}         & 19.6      & 41.2          & 17.4          & 11.7          & 26.0  \\
& \multicolumn{2}{l|}{{\text{GenVIS}}~\cite{heo2022generalized}} &\textcolor{lightgray}{Offline}                         & 34.5      & 59.4          & 35.0          & 16.6        & 38.3  \\ 
\cmidrule{2-9}
& \multicolumn{2}{l|}{CrossVIS~\cite{yang2021crossover}} &Online    & 14.9      & 32.7          & 12.1          & 10.3          & 19.8  \\
& \multicolumn{2}{l|}{VISOLO~\cite{han2022visolo}}    &Online     & 15.3      & 31.0          & 13.8          & 11.1          & 21.7  \\
& \multicolumn{2}{l|}{MinVIS~\cite{huang2022minvis}}   &Online      & 25.0      & 45.5          & 24.0          & 13.9          & 29.7  \\
& \multicolumn{2}{l|}{IDOL~\cite{wu2022defense}}      &Online       & 30.2      & 51.3          & 30.0          & 15.0          & 37.5  \\
& \multicolumn{2}{l|}{{\text{GenVIS}}~\cite{heo2022generalized}}&Online  & \textbf{35.8} & \textbf{60.8}          & \underline{36.2}          & \textbf{16.3}          & \textbf{39.6}   \\ 

& \multicolumn{2}{l|}{\text{TCOVIS}} &Online & \underline{35.3} & \underline{60.7} & \textbf{36.6} & \underline{15.7} & \underline{39.5}  \\

\hline

\multirow{6}{*}{\rotatebox{90}{Swin-L}}
& \multicolumn{2}{l|}{VITA~\cite{heo2022vita}}     &\textcolor{lightgray}{Offline}        & 27.7      & 51.9          & 24.9          & 14.9          & 33.0  \\
& \multicolumn{2}{l|}{{\text{GenVIS}}~\cite{heo2022generalized}}&\textcolor{lightgray}{Offline} & 45.4    & 69.2          & 47.8          & 18.9         & 49.0  \\
\cmidrule{2-9}
& \multicolumn{2}{l|}{MinVIS~\cite{huang2022minvis}}  &Online       & 39.4      & 61.5          & 41.3          & 18.1          & 43.3  \\
& \multicolumn{2}{l|}{IDOL~\cite{wu2022defense}}      &Online       & 42.6      & 65.7          & 45.2          & 17.9          & \underline{49.6}  \\
& \multicolumn{2}{l|}{{\text{GenVIS}}~\cite{heo2022generalized}} &Online& \underline{45.2}      & \underline{69.1}         & \underline{48.4}          & \textbf{19.1}         & 48.6  \\

& \multicolumn{2}{l|}{\text{TCOVIS}} &Online& \textbf{46.7} & \textbf{70.9} & \textbf{49.5} & \textbf{19.1} & \textbf{50.8} \\
\hline
\end{tabular}
}
\vspace{1mm}
\caption{
Quantitative results on \textbf{OVIS validation} set.
We \textbf{bold} the highest accuracy and \underline{underline} the second.
$\dagger$ denotes using MsgShifT~\cite{yang2022temporally} backbone.
}
\vspace{-4mm}
\label{tab:ovis}
\end{table}

\begin{table*}[t]
\centering
{
\begin{tabular}{cc|c|ccccc|ccccc}
\hline
\multirow{2}{*}{LocPro} & \multirow{2}{*}{GIA} & \multirow{2}{*}{STE} & \multicolumn{5}{c|}{YouTube-VIS 2019} & \multicolumn{5}{c}{YouTube-VIS 2021}\\
  &  &  & AP                                & AP$_{50}$     & AP$_{75}$     & AR$_{1}$      & AR$_{10}$ & AP     & AP$_{50}$ & AP$_{75}$     & AR$_{1}$      & AR$_{10}$ \\
\hline
\hline

\checkmark & - & - & 50.6  & 71.5  & 56.1  & 47.6  & 58.2
                & 47.4  & 68.1  & 52.1  & 39.8  & 53.1 \\
\hline
- & \checkmark & - & 51.4	& 72.3	& 57.0	& 49.5	& 59.9
                   & 48.6	& 69.5	& 53.6	& 40.6	& 55.0 \\
- & \checkmark & \checkmark & 52.3	& 73.5	& 57.6	& 49.8	& 60.2
                            & 49.5	& 71.2	& 53.8	& 41.3	& 55.9 \\
\hline
\end{tabular}
}
\vspace{2mm}
\caption{
Ablation study of the method for Global Instance Assignment strategy (GIA) and the Spatio-temporal Enhancement module (STE) on the YouTube-VIS 2019 / 2021 validation sets. LocPro denotes the Local Matching and Propagating technique.
}

\label{tab:ablation_method}
\vspace{-3mm}
\end{table*}

\begin{table}[t]
\centering
{
\begin{tabular}{c|ccccc}
\hline
Architecture  & AP        & AP$_{50}$     & AP$_{75}$     & AR$_{1}$      & AR$_{10}$ \\
\hline
\hline
Cross-Attn.     & 49.5	& 71.2	& 53.8	& 41.3	& 55.9 \\
\hline
Concat.     & 47.9	& 68.6	& 52.7	& 41.4	& 55.1 \\
Self-Cross      & 48.2	& 70.0	& 51.8	& 41.0	& 54.1 \\
Dec. Mod.       & 48.5	& 69.7	& 53.9	& 40.6	& 54.7 \\
\hline
Resp. Pos.       & 48.8	& 70.3	& 52.5	& 41.5	& 55.5 \\
Ins. Attn.      & 49.2	& 70.6	& 53.5	& 40.9	& 55.1 \\
\hline
\end{tabular}
}
\vspace{2mm}
\caption{
Ablation study of the module manipulations of the Spatio-temporal Enhancement module on YouTube-VIS 2021.
}
\label{tab:ste_arch}
\vspace{-4mm}
\end{table}

\textbf{YouTube-VIS 2019\&2021.}
From Table~\ref{tab:ytvis2019_2021}, we can observe that TCOVIS has achieved very competitive performance using both lightweight backbones (ResNet-50) and powerful ones (Swin-L). Moreover, our method can even show better performance than offline methods, such as VITA~\cite{heo2022vita} and offline version GenVIS~\cite{heo2022generalized}. On the more difficult YouTube-VIS 2021 dataset, TCOVIS surpasses GenVIS~\cite{heo2022generalized} in AP not only with ResNet-50 by 2.4 but also with Swin-L by 1.7.

\textbf{YouTube-VIS 2022.}
As shown in Table~\ref{tab:ytvis_2022}, TCOVIS performed best in both AP and AR on YouTube-VIS 2022 dataset, which is more challenging than 2019\&2021 datasets. Especially with powerful backbone Swin-L, TCOVIS outperformed GenVIS~\cite{heo2022generalized} in AP with a huge margin of 4.9, which shows the effectiveness of our method for the complex scenarios.

\textbf{OVIS.}
Table~\ref{tab:ovis} presents the comparisons on OVIS dataset, and we can find that TCOVIS also achieved the best 46.7 AP and  19.1 AR with Swin-L backbone. The results demonstrate that TCOVIS can deal with complicated situations where objects are heavily occluded in others. With ResNet-50 backbone, the performance is still the second best compared to the previous state-of-the-art methods.

\subsection{Ablation Study}
In this section, we provided ablation studies and discuss the effects of different settings in the proposed method. The experiments are conducted with a ResNet-50~\cite{he2016deep} backbone on YouTube-VIS 2019/2021~\cite{yang2019video} valid set.

\noindent\textbf{Effectiveness of the proposed assignment strategy and enhancement module.} The ablation studies on the global instance assignment strategy and the spatio-temporal enhancement module are shown in Table~\ref{tab:ablation_method}. As for the assignment strategy, compared to LocPro as the baseline, the model with the proposed global assignment outperformed the baseline model by more than 0.8 on AP, AP$_{50}$ and AP$_{75}$ on YouTube-VIS 2019, and more than 1.2 on YouTube-VIS 2021. The consistently significant improvement indicates that the strategy with global optimal matching contributes to better overall segmentation, while LocPro only considers the local optimal results. Besides, the baseline forces the posterior features to conform to those at the very beginning leading to temporal inconsistency, since the accumulative error impairs the model during the training process. Our proposed assignment strategy encourages the features of all time to fit the global optimal objective, which effectively enhances the temporal consistency.

As for the effectiveness of the spatio-temporal enhancement module, the comparison is also shown in Table~\ref{tab:ablation_method}. The results show that the enhancement module brings improvements of 0.9 AP on YouTube-VIS 2019\&2021, compared to using the proposed assignment strategy individually. In particular, the proposed module improved the performance by 1.2 and 1.7 in AP$_{50}$ on two datasets, respectively. The performance improvements demonstrate that the proposed module effectively captures the spatial information from the previous frame and aggregates the semantic and spatial features across time. As a result, the delicate spatio-temporal design further enhances the temporal consistency of features for the online video instance segmentation model.

Incorporating the assignment strategy and the enhancement module, our proposed method gained remarkable performance improvements of 1.6 AP on YouTube-VIS 2019 and 2.1 AP on YouTube-VIS 2021 over the LocPro baseline. In a nutshell, the experimental results indicate that the temporally consistent features learned by our proposed TCOVIS significantly boost the performances on online video instance segmentation.

\begin{figure*}[b] 
\centering
\includegraphics[width=0.95\textwidth]{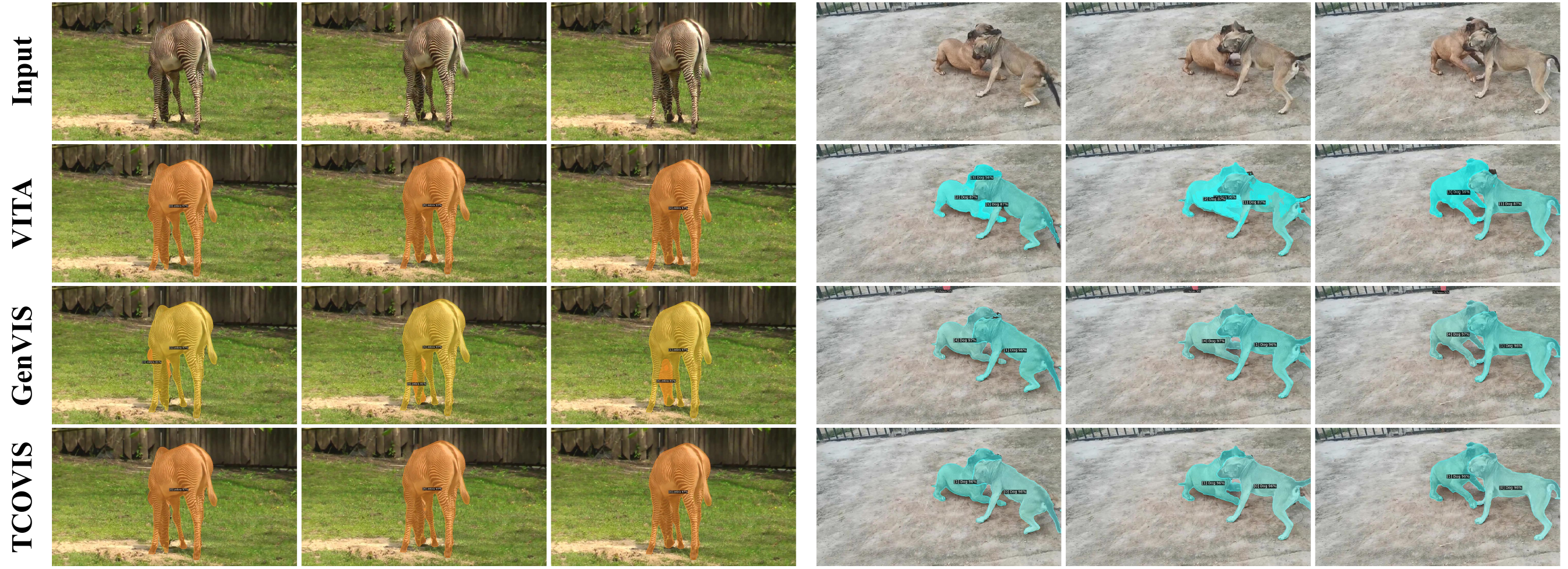}
\caption{Qualitative results of TCOVIS, compared with VITA~\cite{heo2022vita} and GenVIS~\cite{heo2022generalized}. On the left are the predictions on YouTube-VIS 2019~\cite{yang2019video} and on the right are on OVIS~\cite{qi2022occluded}. Objects displayed in the same color denote the same instance. Our TCOVIS shows more temporally consistent results in these challenging scenes, where there are occlusions of the instance itself or others (Best viewed in color).
} 
\label{fig:visualization}
\end{figure*}

\noindent\textbf{Different manipulations of spatio-temporal enhancement.}
In Table ~\ref{tab:ste_arch}, we compared our proposed spatio-temporal enhancement module with other optional manipulations. \emph{Cross-Attn.} denotes our proposed manipulation following the standard multi-head cross-attention layer to decode the spatial features with shared positional embedding described in Section~\ref{sec:ste}. \emph{Concat.} indicates that we concatenated the corresponding spatial feature and prototype of an instance followed by an MLP to get the updated query. \emph{Self-Cross} stands for adding an extra self-attention layer ahead of the cross-attention layer. \emph{Dec. Mod.} is decoder modulation, reviewing the spatial feature for every Transformer decoder layer. \emph{Resp .Pos.} denotes respective positional embeddings were used for the spatial feature and prototype when we performed cross-attention. \emph{Ins. Attn.} indicates the instance-wise decoding in cross-attention layer. 

As shown, compared to all its counterparts, our proposed manipulation achieved the best performance. The first three variants are related to the feature aggregation, where \emph{Concat.} is too naive to model the spatial and semantic features, while \emph{Self-Cross} obscures the spatial feature. We inferred the performance decrease of \emph{Dec. Mod.} comes from the information redundancy when aggregating them in every layer. The last two experiments studied the instance-wise correspondence. In the \emph{Resp. Pos.} setting, explicit correspondence for spatial and semantic features of the same instance between frames is absent. \emph{Ins. Attn.} only focuses on the instance itself across time and can slightly enhance the temporal consistency, however, neglecting the spatial features of other instances, the module fails to capture the spatial relationship among instances. The results confirm that the proposed manipulation effectively exploits the spatial information along the temporal dimension to enhance the temporal consistency of features.

\begin{figure*}[t] 
\centering
\includegraphics[width=0.95\linewidth]{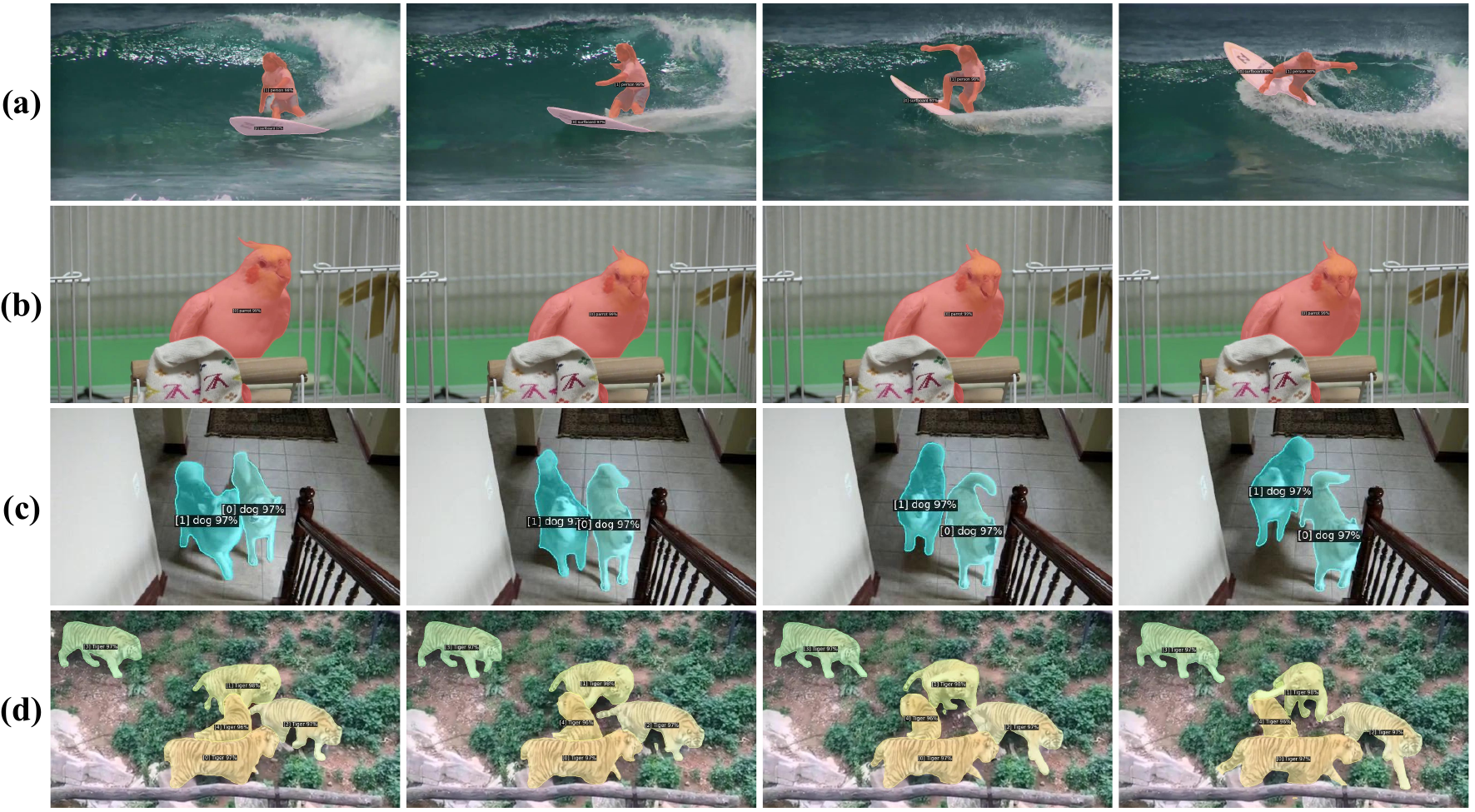}
\vspace{1mm}
\caption{Qualitative results of TCOVIS in four challenging  cases on YouTube-VIS 2019~\cite{yang2019video} and OVIS~\cite{qi2022occluded}: (a) fast movement, (b) heavy occlusion, (c) low resolution, and (d) crowded scene. Objects displayed in the same color denote the same instance. The qualitative results demonstrate the effectiveness and robustness of TCOVIS (Best viewed in color).}
\label{fig:supp_visual}
\vspace{-2mm}
\end{figure*}

\subsection{Qualitative results}
In Figure~\ref{fig:visualization}, we show the qualitative comparisons of the proposed TCOVIS with VITA~\cite{heo2022vita} and GenVIS~\cite{heo2022generalized} on YouTube-VIS 2019 and OVIS datasets. In the left scene where there is a zebra with self-occlusion, GenVIS fails to segment its head resulting in fragmented predictions, however, TCOVIS performs temporally consistent segmentation with impressive accuracy. On the right is a difficult case where there are two dogs with similar appearances grappling with each other. VITA incorrectly detects more than two dogs and fails to track the left one, while GenVIS fails to handle the margin of two instances, e.g. the nose and the paw of the left dog. TCOVIS successfully tracks and segments the instances, demonstrating its effectiveness.

In Figure~\ref{fig:supp_visual}, we provide more qualitative results of the proposed method in variously challenging  cases, which are all chosen from the mentioned benchmarks~\cite{yang2019video,qi2022occluded}. As shown in Figure~\ref{fig:supp_visual} (a), our method successfully tracks the surfer and the surfboard with fast movement. In the second row, we present a difficult case with severe occlusion, in which our method performs admirably by accurately segmenting the claw (at the bottom) of a parrot despite the presence of a wooden stick that partially obstructs its lower body. Figure~\ref{fig:supp_visual} (c) illustrates a low-resolution case and our method still achieves good performance. In Figure~\ref{fig:supp_visual} (d), we depict a crowded scene where TCOVIS is capable of handling multiple instances that share similar appearances and exhibit complex interactions. All the qualitative results in the challenging situations demonstrate the effectiveness and robustness of the proposed method.

\section{Conclusion}
In this paper, we propose a new online video instance segmentation method, TCOVIS, to fully exploit the temporal information within a video and produce temporally consistent predictions. Based on the query propagation framework, we propose a global instance assignment strategy to perform global optimal matching with the consideration of the entire video and supervise the model with the global optimal objective. We further devise a spatio-temporal enhancement module to capture the spatial feature and aggregate it with the semantic feature between frames. The effectiveness of our method is evaluated with experimental results on YouTube-VIS 2019/2021/2022 and OVIS.

\section*{Acknowledgement}

This work was supported in part by the National Key Research and Development Program of China under Grant 2022ZD0160102 and in part by the National Natural Science Foundation of China under Grant 62125603.

{\small
\bibliographystyle{ieee_fullname}
\bibliography{egbib}
}

\end{document}